\pdfoutput=1

\documentclass[11pt]{article}

\usepackage[]{EMNLP2023}

\usepackage{times}
\usepackage{latexsym}

\usepackage[T1]{fontenc}

\usepackage[utf8]{inputenc}

\usepackage{microtype}

\usepackage{inconsolata}

\usepackage{booktabs}
\usepackage{tabularx}
\usepackage{graphicx}
\usepackage{multirow}
\usepackage{amsmath}

\usepackage{tcolorbox}                    
\tcbuselibrary{listings, breakable}       

\newtcolorbox{promptbox}[1][]{           
  colback=gray!5,
  colframe=gray!50,
  fonttitle=\bfseries\small,
  title=#1,
  breakable,
  left=6pt, right=6pt,
  top=4pt, bottom=4pt,
  fontupper=\small\ttfamily
}

\title{ZAS-SQL: Distilling Rules from Failures for Zero-Shot Text-to-SQL}

\author{
    \textbf{Hongzhou Zheng}$^{1,3}$, \textbf{Yixin Gou}$^{1,3}$, \textbf{Wenjia Zhang}$^{1,2,3}$ \\
    $^1$Shanghai Research Institute for Intelligent Autonomous Systems, Tongji University \\
    $^2$College of Architecture and Urban Planning, Tongji University \\
    $^3$Behavioral and Spatial AI Lab, Peking University \& Tongji University \\
    \texttt{hongzhouzheng@tongji.edu.cn, wenjiazhang@tongji.edu.cn}
    }

\begin{document}
\maketitle

\footnotetext[1]{Appendices are omitted and will be included in the camera-ready version.}

\begin{abstract}
Text-to-SQL translates natural language into executable SQL queries. Few-shot in-context learning methods built upon large language models (LLMs) achieve strong performance, yet their reliance on demonstrations limits cross-domain generalization and consumes substantial context window space. Existing zero-shot methods, lacking effective generation constraints, still fall short of few-shot approaches.
We observe that LLM failures in zero-shot Text-to-SQL are not random but exhibit systematic, recurring patterns. Building on this observation, we propose a fully zero-shot Text-to-SQL framework that distills core generation rules from failure cases through a Map-Reduce-based rule distillation pipeline and improves generation quality via three complementary modules: knowledge-augmented schema representation, which supplements missing semantics in Data Definition Language; a rule-driven structured reasoning framework that suppresses structural deviations; and Execution-Guided Early Stopping, which enables low-cost self-correction.
On Spider, the proposed framework achieves up to 87.2\% and 88.6\% execution accuracy on the Dev and Test sets, respectively, establishing a new zero-shot state-of-the-art and surpassing multiple few-shot and fine-tuning methods built upon GPT-4/4o. On the domain-specific dataset UrbanPlan, it achieves 81.3\%, confirming that the rule distillation approach generalizes across domains. Moreover, when equipped with a 4B-parameter model, the framework surpasses zero-shot baselines of leading closed-source models, demonstrating strong model generality.
\end{abstract}

\section{Introduction}\label{Introduction}
Text-to-SQL~\citep{1-katsogiannis-meimarakis_survey_2023, 2-deng_recent_nodate, 3-hong_next-generation_2025} translates natural language into executable SQL queries, enabling non-expert users to access relational databases with significant practical value in data analysis~\citep{4-huang_exploring_2025}. With the advancement of large language models (LLMs)~\citep{37-JMLR:v21:20-074,38-lewis-etal-2020-bart,39-chen2021evaluatinglargelanguagemodels,41-openai2022chatgpt,6-openai2024gpt4technicalreport,40-openai-gpt4o}, the field has achieved notable progress on benchmarks such as Spider~\citep{5-yu_spider_2019, 6-openai2024gpt4technicalreport, 7-shi_survey_2018, 29-NEURIPS2023_83fc8fab}. However, existing methods generally rely on task-specific supervision signals, including annotated data for fine-tuning and curated demonstrations for few-shot learning, severely limiting their practicality in cold-start and cross-domain deployment scenarios.

Existing methods fall into two main categories. Fine-tuning methods improve SQL generation through supervised training~\citep{8-gorti_msc-sql_nodate, 9s-zhong-etal-2024-learning, 10-zhang_finsql_2024, 11s-yang-etal-2024-synthesizing, 12s-pourreza-rafiei-2024-dts}, but suffer from high computational cost, heavy data dependence, and limited cross-domain generalization. Few-shot in-context learning (ICL) methods guide generation through fixed or dynamically retrieved demonstrations~\citep{13s-zhang-etal-2023-act,14-wang_mac-sql_nodate,15s-sun-etal-2023-sqlprompt,16-pourreza_din-sql_nodate, 17-gao_text--sql_2023}, reducing training cost yet still facing constrained generalization, retrieval failures, and excessive context consumption that limits the space available for schema representation in complex multi-table scenarios. These limitations hinder the deployment of Text-to-SQL in vertical domains such as urban planning, where annotated corpora are scarce and database schemas evolve frequently.

To address deployment challenges, we propose ZAS-SQL, a fully zero-shot Text-to-SQL framework. Our key observation is that LLM failures in zero-shot Text-to-SQL are not random but exhibit systematic, recurring patterns. The main contributions are as follows.

\paragraph{Transferable Rule Distillation.} We design a Map-Reduce-based automated pipeline that extracts domain-adapted generation rules from zero-shot failure cases through error diagnosis and pattern aggregation. Experiments on Spider, a widely used public benchmark, and UrbanPlan, a dataset of real SQL tasks collected from daily workflows at a professional planning institute, show that the pipeline consistently distills high-gain rules, confirming cross-domain generalizability.

\paragraph{Knowledge-Augmented Schema Representation.} We introduce an automated semantic annotation mechanism that, through sampled value extraction and LLM-based annotation, supplements missing business semantics and value distributions in DDL. This module requires zero manual annotation and incurs no additional inference overhead, reducing schema linking errors.

\paragraph{Execution-Guided Early Stopping (EGES).} We propose a self-correction strategy that performs dynamic early stopping based on execution result consistency. This substantially reduces inference cost without requiring sequential iterative refinement.

\paragraph{Strong Empirical Results.} On Spider, ZAS-SQL powered by DeepSeek-V3 achieves 87.2\%/87.8\% (Dev/Test) zero-shot execution accuracy, to our knowledge the strongest zero-shot result to date, surpassing multiple few-shot and fine-tuning methods. With a 4B-parameter model, the framework still reaches 86.3\% (Test), confirming generality across model scales.

\section{Related Work}\label{Related Work}
\paragraph{Fine-tuning and Few-shot Learning.} Recent progress in Text-to-SQL has largely relied on LLM fine-tuning and few-shot in-context learning (ICL)~\citep{44-lee-etal-2025-safe-sql}. Fine-tuning methods~\citep{11s-yang-etal-2024-synthesizing, 12s-pourreza-rafiei-2024-dts,42-ICLR2025_212b143b} use synthetic data and preference alignment to bring open-source models to the level of GPT-4~\citep{6-openai2024gpt4technicalreport}, but face bottlenecks including high training cost, heavy data dependence, and limited cross-domain generalization. Few-shot ICL guides generation through elaborate prompt engineering~\citep{16-pourreza_din-sql_nodate, 35-NEURIPS2022_9d560961} or dynamic demonstration  retrieval~\citep{17-gao_text--sql_2023,18-lee-etal-2025-mcs}, yet demonstrations heavily consume the limited context window and retrieval failures can readily mislead the model. These constraints motivate exploring a low-cost, generalizable alternative under a fully zero-shot setting.

\paragraph{Sampling Consistency and Execution Feedback.} To reduce the randomness inherent in LLM generation, most studies adopt self-consistency voting~\citep{18-lee-etal-2025-mcs,19-wang_self-consistency_2023,20-chen_codet_2022,21-ICLR2025_974ff7b5} or execution feedback~\citep{14-wang_mac-sql_nodate}. DART-SQL~\citep{22-mao-etal-2024-enhancing} and MAGIC~\citep{23-Askari_Poelitz_Tang_2025} employ sequential iterative refinement via self-debugging. However, fixed-round multi-path sampling incurs prohibitive inference overhead, and sequential feedback often triggers error cascading in complex queries due to uncontrollable iteration depth. Inspired by~\citet{24-aggarwal-etal-2023-lets}, our EGES mechanism abandons sequential iteration and instead performs dynamic early stopping through parallel sampling and execution result set consistency. Unlike CHASE-SQL~\citep{21-ICLR2025_974ff7b5}, which relies on fixed sampling coupled with an offline-trained selector, EGES requires no additional training and substantially reduces inference cost while preventing error propagation.

\paragraph{Zero-shot Generation and Schema Augmentation.} \citet{25-dong_c3_2023} and \citet{43-li2025alphasqlzeroshottexttosqlusing} have demonstrated zero-shot viability, yet performance degrades sharply on real-world databases containing abbreviations or domain-specific terminology, where Data Definition Language (DDL) semantics alone prove insufficient. Traditional schema linking relies on surface string matching~\citep{26-gan-etal-2023-appraising} and fails frequently. Recent prompt augmentation approaches face a dilemma: injecting database content~\citep{27-shen-etal-2024-improving} introduces data privacy risks and computational overhead, while manually annotated data dictionaries~\citep{28-lee-etal-2021-kaggledbqa,29-NEURIPS2023_83fc8fab} cannot scale. Our approach instead caches business semantics inferred by an LLM offline, achieving schema augmentation with zero manual annotation and zero additional inference overhead.

\section{Error Analysis and Rule Distillation.}\label{Error Analysis and Rule Distillation}
To identify the core failure modes of LLMs in zero-shot Text-to-SQL, we establish a zero-shot baseline on Spider Train (using DeepSeek-V3~\citep{30-deepseekai2025deepseekv3technicalreport} with only the natural language question and DDL-formatted schema as input~\citep{17-gao_text--sql_2023,31-nan-etal-2023-enhancing}), collecting 1,725 failure cases (overall execution accuracy: 80.1\%).

\subsection{Scalable Rule Distillation}\label{Scalable Rule Distillation}
Manual error analysis~\citep{16-pourreza_din-sql_nodate} faces inherent limitations in scalability and annotator bias when confronted with thousands of failures. To address these issues, we propose a Map-Reduce-based rule distillation pipeline with Human-in-the-Loop (HITL), executed in three stages:

\paragraph{Local Pattern Extraction.} The LLM analyzes each case along four dimensions: error type, reasoning level, failure cause, and missing information, producing a structured error report per case. This stage yields 1,722 reports (3 cases excluded due to format parsing failures).

\paragraph{Global Pattern Aggregation.} Local reports are aggregated and deduplicated through LLM-based abstraction, producing 20 mutually independent global error patterns.

\paragraph{Retrospective Alignment.} The 20 global patterns re-label all reports, eliminating annotation drift and aligning 1,715 samples (alignment rate 99.6\%).

\medskip
\noindent
Statistical results show that the top-10 patterns cover 81\% of all failures, which can be grouped into three core challenges. Over-complication: generating redundant JOINs or subqueries that violate SQL simplicity. Schema semantic deficiency: the absence of business semantics and value distributions in DDL causes schema linking errors. Single-pass reasoning fragility: complex set operations and implicit semantic mappings fail under one-pass autoregressive decoding.

\subsection{Core Rule Derivation}\label{Core Rule Derivation}
\begin{figure*}[t]
    \centering
    \includegraphics[width=\textwidth]{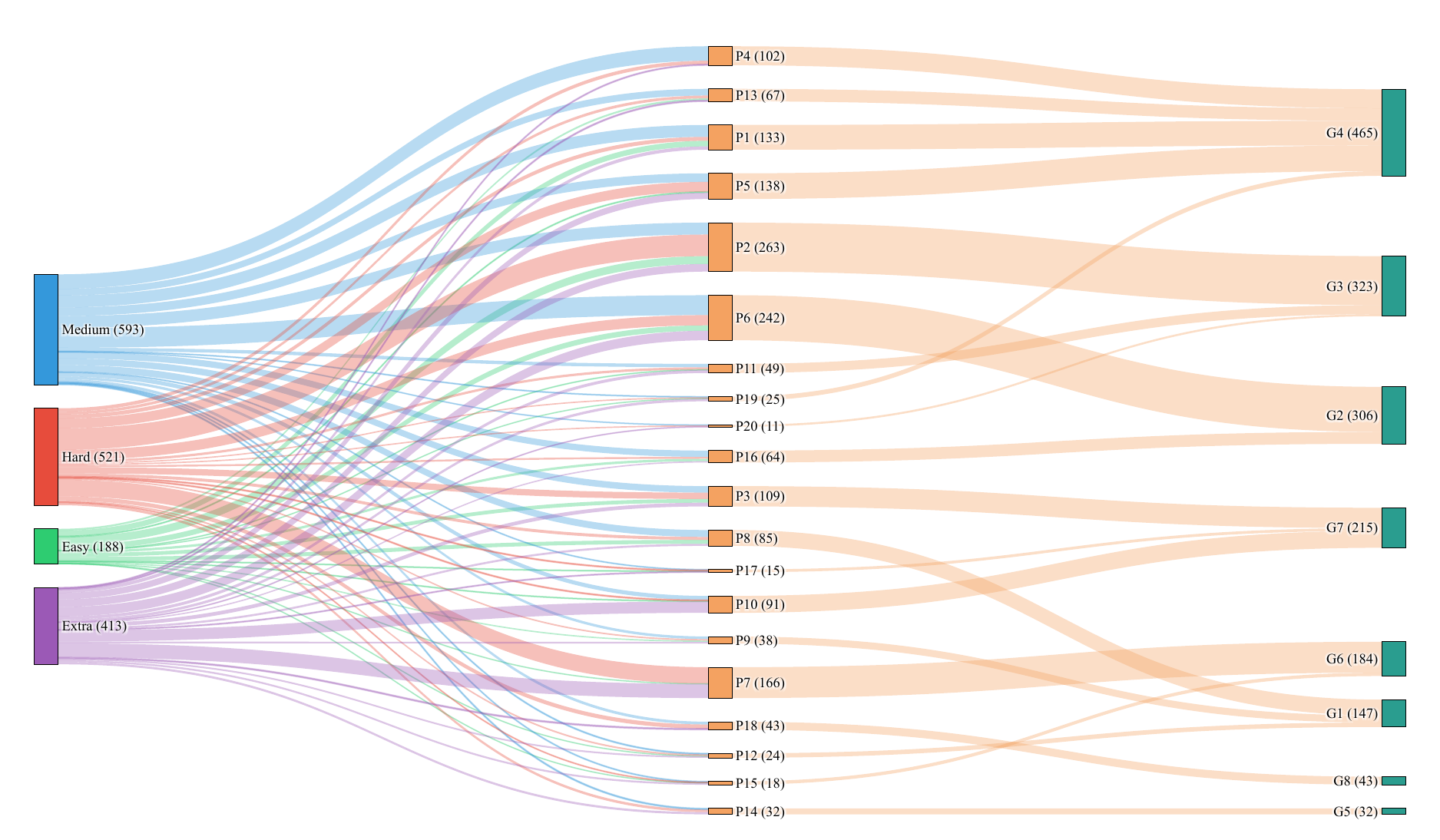}
    \caption{Sankey diagram of the three-stage rule distillation pipeline. Left: 1,725 zero-shot failure cases from Spider Train, stratified by difficulty level. Center: 20 global error patterns produced by Map-Reduce aggregation. Right: 8 actionable core generation rules derived through expert abstraction. Flow widths are proportional to case counts, showing full coverage of the failure corpus by the final rule set.}
    \label{sankey_diagram}
\end{figure*}

The extracted error patterns remain at the failure characterization level and cannot directly serve as generation guidance. We therefore introduce the HITL mechanism for transformation: two domain experts review and abstract the 20 global patterns into 8 actionable core generation rules: R1 value matching and precision, R2 column semantic disambiguation, R3 column selection and output ordering, R4 JOIN strategy, R5 sorting and comparison, R6 set operations, R7 aggregation and correlation strategy, and R8 SQLite data types. As shown in Figure~\ref{sankey_diagram}, all 1,715 samples converge to these 8 rules, demonstrating complete coverage. These rules collectively address the three challenges identified above and form the design foundation of ZAS-SQL. The rule distillation pipeline uses Spider Train for demonstration; when applied to a new domain, the same process can be reused directly to distill domain-adapted rules.

The HITL step operates on aggregated patterns rather than individual cases and constitutes a one-time, lightweight effort. Because Text-to-SQL tasks share fundamental structural properties (schema linking, JOIN reasoning, aggregation logic), the resulting rules are inherently reusable across domains; moreover, when the same pipeline is applied to UrbanPlan with only 150 failure cases, it still yields substantial accuracy gains (Section~\ref{Ablation Study}), demonstrating that the distillation process itself transfers to new domains.

\section{Methodology}\label{Methodology}
\begin{figure*}[t]
    \centering
    \includegraphics[width=\textwidth]{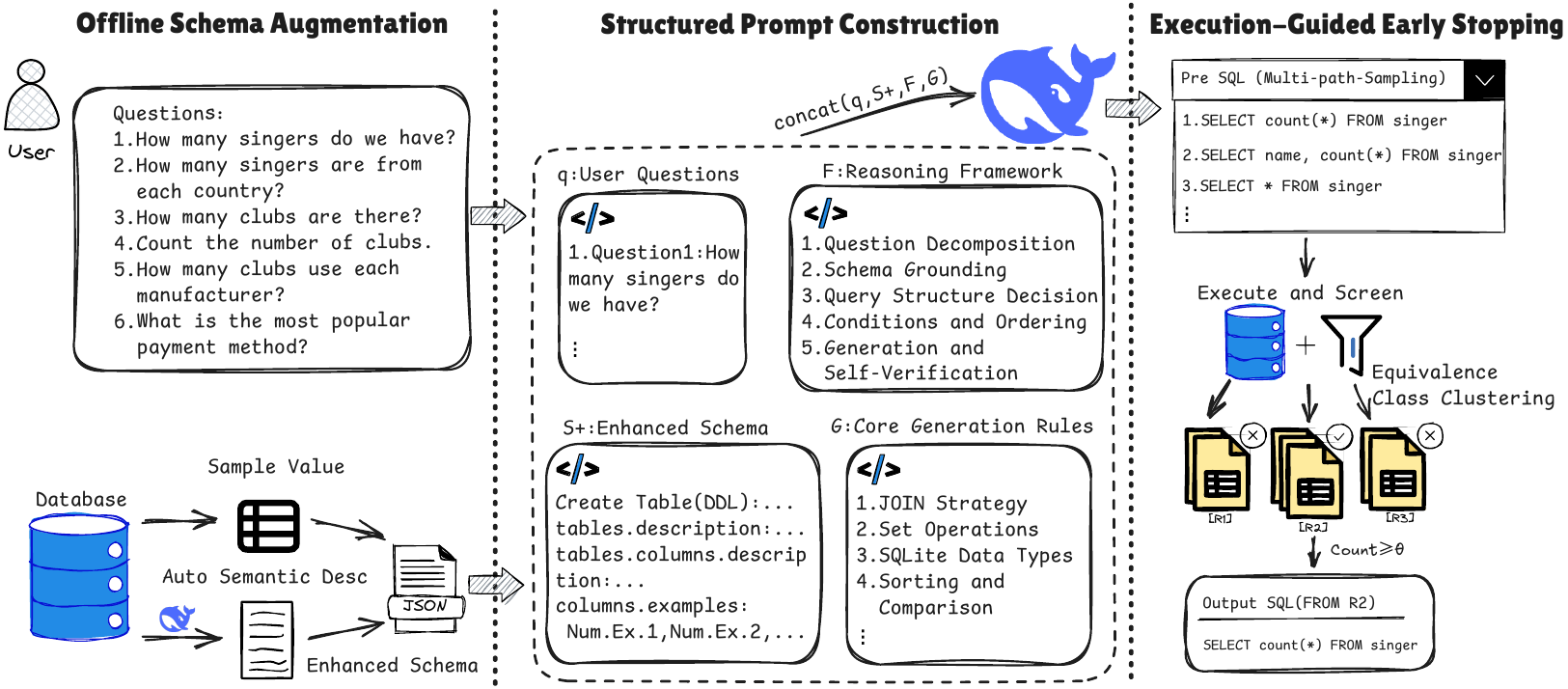}
    \caption{Overall architecture of ZAS-SQL. (a) Offline schema augmentation: sampled values and LLM-inferred semantic descriptions are appended to raw DDL to form the knowledge-augmented schema $\mathcal{S}^+$. (b) Structured prompt construction: the natural language question $q$, augmented schema $\mathcal{S}^+$, core generation rules $\mathcal{G}$, and five-step reasoning framework $\mathcal{F}$ are assembled into a unified prompt $\mathcal{P}$. (c) Execution-Guided Early Stopping (EGES): candidate SQL queries sampled in parallel are executed against the database and grouped into equivalence classes by result set; sampling terminates once any class reaches the consistency threshold $\theta$.}
    \label{text2sql}
\end{figure*}

ZAS-SQL operates under a fully zero-shot setting, guiding an LLM to translate a natural language question $q$ into an executable SQL query $y^*$ solely through prompting. To address the three core challenges identified in Section~\ref{Error Analysis and Rule Distillation}, the framework comprises three cooperative modules (Figure~\ref{text2sql}): knowledge-augmented schema representation $\mathcal{S}^+$ addresses schema semantic deficiency, the structured reasoning framework addresses over-complication, and EGES addresses single-pass generation fragility. The overall structured prompt is formalized as:
\begin{equation}
\mathcal{P} = \mathrm{Concat}(\mathcal{G},\; \mathcal{S}^+,\; q,\; \mathcal{F})
\end{equation}
where $\mathcal{G}$ denotes the core rules, $\mathcal{S}^+$ the knowledge-augmented schema representation, $q$ the natural language question, and $\mathcal{F}$ the reasoning framework. The LLM performs multi-path sampling with prompt $\mathcal{P}$, and EGES selects the final output.

\subsection{Knowledge-Augmented Schema}\label{Knowledge-Augmented Schema}
To address schema linking errors caused by the absence of business semantics and value distributions in DDL, we construct the augmented representation $\mathcal{S}^+$ through offline preprocessing.

\paragraph{Sample Value Extraction.} For each column, we randomly sample 10 deduplicated values and append them to the DDL, providing the model with a distributional prior that reduces value matching errors (e.g., case insensitivity, fuzzy matching). This quantity balances sufficient value coverage against prompt length: too few samples fail to cover representative values, while too many consume the context window and introduce redundancy.

\paragraph{Automated Semantic Description.} Using column naming conventions and database context, an LLM automatically infers natural language semantic descriptions for tables and columns. Compared to manually constructed data dictionaries, this method offers plug-and-play applicability and offline preprocessing with no runtime cost.

\medskip
\noindent
The original DDL, semantic descriptions, and sample values are concatenated to form $\mathcal{S}^+$, bridging the semantic gap between natural language and the database schema.

\subsection{Structured Reasoning Framework}\label{Structured Reasoning Framework}
To suppress the over-complication tendency of zero-shot LLMs in SQL construction, we explicitly inject the core generation rules $\mathcal{G}$ from Section~\ref{Error Analysis and Rule Distillation} and the reasoning framework $\mathcal{F}$ into the prompt. Compared to few-shot methods that rely on implicit learning from demonstrations, explicit rule constraints offer stronger cross-domain generalizability.

The reasoning framework $\mathcal{F}$ enforces step-by-step reasoning: (1) intent decomposition, analyzing the core query intent; (2) schema grounding, identifying relevant tables and columns; (3) structural decision, determining SQL structures such as JOINs and subqueries; (4) condition and ordering logic; and (5) generation and self-verification, outputting SQL and performing self-checks against the rules. Each step embeds inspection directives targeting high-frequency error patterns, effectively constraining the generation boundary.

\subsection{Execution-Guided Early Stopping}\label{Execution-Guided Early Stopping}
To reduce single-pass generation randomness (multi-path sampling serves as inference-time variance reduction) and avoid the token waste imposed by fixed sampling counts in conventional self-consistency~\citep{19-wang_self-consistency_2023}, we propose EGES. This mechanism evaluates consistency at the level of SQL execution result sets rather than SQL text, correctly handling semantically equivalent SQL variants that differ only in surface form.

Given question $q$, database $\mathcal{D}$, prompt $\mathcal{P}$, sampling temperature $\tau$, and consistency threshold $\theta$, candidate $s_i$ from the $i$-th sample is executed on $\mathcal{D}$ to obtain result $R_i = \mathrm{Exec}(s_i, \mathcal{D})$. The equivalence class and final output are defined as:
\begin{align}
C_R &= \{s_i \mid \mathrm{Exec}(s_i, \mathcal{D}) = R,\; R \neq err\} \\
y^* &= \mathrm{EGES}(\mathcal{P}, \theta) = \arg\min_{s \,\in\, C^*} |s|, \;\; |C^*| \geq \theta
\end{align}
where $s_i$ denotes the $i$-th candidate SQL query, $C_R$ is the equivalence class of candidates sharing execution result $R$, and $C^*$ is the largest class reaching threshold $\theta$.

If $s_i$ triggers a syntax or execution error, it is discarded and excluded from the effective sample count; a legitimate empty result set $R_i = \emptyset$ participates in consistency statistics normally. Once any equivalence class reaches cardinality $\theta$, sampling terminates immediately. If the upper bound $N_{\max}$ is reached without triggering early stopping, the shortest SQL query by character length is selected from the winning equivalence class $C^*$. Since all candidates within the same class share identical execution result sets, this selection does not affect execution accuracy while favoring readability.

\section{Experiments}\label{Experiments}
\subsection{Datasets and Evaluation Metrics}\label{Datasets and Evaluation Metrics}
\paragraph{Spider.} Spider is one of the most widely used cross-domain Text-to-SQL benchmarks~\citep{5-yu_spider_2019}. Its training set (8,659 examples, 146 databases), development set (1,034 examples, 20 databases), and test set (2,147 examples, 34 databases) share no database overlap, comprising 11,840 question-SQL pairs across 200 databases spanning 138 domains. Queries are categorized into four difficulty levels: Easy, Medium, Hard, and Extra.

\paragraph{UrbanPlan.} UrbanPlan is a domain-specific Text-to-SQL dataset for urban planning, derived from the operational databases and real-world analytical tasks of a municipal planning institute. It contains 375 question-SQL pairs covering 3 sub-databases and 19 tables, spanning scenarios such as population mobility, commuting patterns, demographic profiling, and housing markets. Because both the relational schemas and the analytical questions originate from routine professional workflows, the dataset captures the complexity and diversity characteristic of production-level planning databases. Of these, 150 samples are used for rule distillation and 225 for evaluation, providing a vertical-domain complement to the broad cross-domain coverage of Spider.

Queries are classified into five complexity levels based on SQL structural characteristics, ranging from single-table queries to multi-table JOINs and nested subqueries.

\paragraph{Evaluation Metrics.} We adopt execution accuracy (EX) as the primary metric. Unlike exact set match, EX compares execution result sets and thus accommodates semantically equivalent SQL variants with different structures, making it the standard metric for assessing practical usability of LLM-based methods. We use the official Spider evaluation suite for all measurements.

\subsection{Baselines and Experimental Setup}\label{Baselines}
\paragraph{Zero-shot Methods.} ChatGPT-SQL~\citep{32-liu2023comprehensiveevaluationchatgptszeroshot}, C3-SQL~\citep{25-dong_c3_2023}, and Alpha-SQL~\citep{43-li2025alphasqlzeroshottexttosqlusing}.

\paragraph{Few-shot ICL Methods.} DIN-SQL~\citep{16-pourreza_din-sql_nodate}, DAIL-SQL~\citep{17-gao_text--sql_2023}, ACT-SQL~\citep{13s-zhang-etal-2023-act}, DEA-SQL~\citep{33-xie-etal-2024-decomposition}, MCS-SQL~\citep{18-lee-etal-2025-mcs}, OpenSearch-SQL~\citep{34-10.1145/3725331}, and MAC-SQL~\citep{14-wang_mac-sql_nodate}.

\paragraph{Fine-tuning Methods.} SENSE-SQL~\citep{11s-yang-etal-2024-synthesizing}, DTS-SQL~\citep{12s-pourreza-rafiei-2024-dts}, and ROUTE-SQL~\citep{42-ICLR2025_212b143b}.

\paragraph{Experimental Setup.} Since several baselines rely on commercial APIs (e.g., GPT-4/4o), we select DeepSeek-V3~\citep{30-deepseekai2025deepseekv3technicalreport} as the primary backbone for its strong zero-shot capability and open-source reproducibility. To isolate the effect of backbone differences, we assess framework contributions through multi-model benchmarks and relative improvement margins. Following~\citet{17-gao_text--sql_2023}, we disable thinking mode during experiments. EGES parameters are set uniformly: sampling temperature $\tau = 1$, maximum sampling count $N_{\max} = 20$, and early stopping threshold $\theta = 6$. GPT-4o~\citep{6-openai2024gpt4technicalreport}, DeepSeek-V3~\citep{30-deepseekai2025deepseekv3technicalreport}, GLM-5~\citep{48-glm5team2026glm5vibecodingagentic}, Kimi-K2.5~\citep{49-kimiteam2026kimik2openagentic} and Qwen3.6-plus~\citep{45-yang2025qwen3technicalreport} are accessed via official APIs; all other models are deployed locally on 8$\times$A800 GPUs.

\subsection{Experiment Results}\label{Experiment Results}
Table~\ref{experiment_results} reports the execution accuracy of ZAS-SQL and representative baselines on Spider Dev and Spider Test.

\begin{table*}[t]
    \centering
    \caption{Execution accuracy (\%) on Spider Dev/Test. Methods are grouped by paradigm: zero-shot, few-shot ICL, and supervised fine-tuning. All ZAS-SQL variants operate under the zero-shot setting with $\theta=6$ and $\tau=1$. (-) indicates values not reported in the original publication. Underlined values indicate the best result under the zero-shot setting.}
    \label{experiment_results}
    \resizebox{\textwidth}{!}{%
    \begin{tabular}{l c c c c c c c c c c}
        \toprule
        \multicolumn{1}{c}{\textbf{Method}} & \multicolumn{5}{c}{\textbf{Dev (EX \%)}} & \multicolumn{5}{c}{\textbf{Test (EX \%)}} \\
        \cmidrule(lr){2-6} \cmidrule(lr){7-11}
         & \textbf{Easy} & \textbf{Medium} & \textbf{Hard} & \textbf{Extra} & \textbf{All} & \textbf{Easy} & \textbf{Medium} & \textbf{Hard} & \textbf{Extra} & \textbf{All} \\
        \midrule
        \multicolumn{11}{c}{\textbf{Zero-shot}} \\
        \midrule
        ChatGPT-SQL + ChatGPT~\citep{32-liu2023comprehensiveevaluationchatgptszeroshot} & - & - & - & - & 72.3 & - & - & - & - & - \\
        C3-SQL + ChatGPT~\citep{25-dong_c3_2023} & 92.7 & 85.5 & 77.6 & 62.0 & 81.8 & - & - & - & - & 82.3 \\
        Alpha-SQL + Qwen2.5-Coder-14B~\citep{43-li2025alphasqlzeroshottexttosqlusing} & 94.0 & 91.0 & 79.9 & \underline{72.3} & 87.0 & - & - & - & - & - \\
        \midrule
        \multicolumn{11}{c}{\textbf{Few-shot}} \\
        \midrule
        DIN-SQL + GPT-4~\citep{16-pourreza_din-sql_nodate} & 92.3 & 87.4 & 76.4 & 62.7 & 82.8 & - & - & - & - & 85.3 \\
        DAIL-SQL + GPT-4~\citep{17-gao_text--sql_2023} & 91.9 & 90.1 & 75.2 & 63.8 & 83.6 & - & - & - & - & 86.6 \\
        ACT-SQL + GPT-4~\citep{13s-zhang-etal-2023-act} & - & - & - & - & 82.9 & - & - & - & - & - \\
        DEA-SQL + GPT-4~\citep{33-xie-etal-2024-decomposition} & 88.7 & 89.5 & 85.6 & 70.5 & 85.6 & - & - & - & - & 87.1 \\
        MCS-SQL + GPT-4~\citep{18-lee-etal-2025-mcs} & 94.0 & 93.5 & 88.5 & 72.9 & 89.5 & - & - & - & - & 89.6 \\
        MAC-SQL + GPT-4~\citep{14-wang_mac-sql_nodate} & - & - & - & - & 86.8 & - & - & - & - & 82.8 \\
        ROUTE(MCP) + Qwen2.5-14B~\citep{42-ICLR2025_212b143b} & - & - & - & - & 80.0 & - & - & - & - & 80.6 \\
        OpenSearch-SQL + GPT-4o\citep{34-10.1145/3725331} & - & - & - & - & - & - & - & - & - & 87.1 \\
        \midrule
        \multicolumn{11}{c}{\textbf{Supervised Fine-Tuning}} \\
        \midrule
        SENSE + CodeLLaMA-13B\citep{11s-yang-etal-2024-synthesizing} & - & - & - & - & 84.1 & - & - & - & - & 86.6 \\
        DTS-SQL + DS-7B\citep{12s-pourreza-rafiei-2024-dts} & - & - & - & - & 85.5 & - & - & - & - & - \\
        ROUTE + Qwen2.5-14B~\citep{42-ICLR2025_212b143b} & 94.0 & 93.0 & 81.6 & 68.1 & 87.3 & - & - & - & - & 87.1 \\
        \midrule
        \multicolumn{11}{c}{\textbf{Ours (Zero-shot)}} \\
        \midrule
        ZAS-SQL + Qwen2.5-3B-Instruct & 86.3 & 74.0 & 58.6 & 45.2 & 69.7 & 87.5 & 75.3 & 54.5 & 52.9 & 69.7 \\
        ZAS-SQL + Qwen2.5-14B-Instruct & 91.9 & 91.7 & \underline{84.5} & 66.9 & 86.6 & 91.9 & 89.9 & 85.8 & 79.0 & 87.6 \\
        ZAS-SQL + Qwen2.5-Coder-14B & \underline{94.3} & \underline{92.4} & 83.9 & 66.3 & \underline{87.2} & 92.3 & \underline{91.0} & \underline{84.9} & 82.6 & \underline{88.6} \\
        ZAS-SQL + Qwen3-4B & 92.3 & 91.5 & 83.9 & 65.1 & 86.2 & \underline{92.8} & 89.0 & 81.6 & 77.3 & 86.3 \\
        ZAS-SQL + Qwen3-8B & 92.3 & 90.1 & 78.7 & 66.3 & 84.9 & 92.6 & 89.6 & 83.8 & \underline{82.9} & 87.9 \\
        ZAS-SQL + DeepSeek-V3 & 93.6 & 92.2 & 81.6 & 69.9 & \underline{87.2} & 92.3 & 90.3 & 84.5 & 79.8 & 87.8 \\
        \bottomrule
    \end{tabular}%
    }
\end{table*}

\begin{table}[t]
    \centering
    \caption{Execution accuracy (\%) of ZAS-SQL on the domain-specific UrbanPlan benchmark.}
    \label{urbanplan_results}
    \begin{tabular*}{0.95\linewidth}{@{\extracolsep{\fill}} lc}
        \toprule
        \textbf{Model} & \textbf{EX (\%)} \\
        \midrule
        ZAS-SQL + Qwen2.5-3B-Instruct         & 45.3 \\
        ZAS-SQL + Qwen2.5-14B-Instruct        & 72.0 \\
        ZAS-SQL + Qwen2.5-Coder-14B        & 68.9 \\
        ZAS-SQL + Qwen3-4B           & 68.0 \\
        ZAS-SQL + Qwen3-8B           & 75.6 \\
        ZAS-SQL + DeepSeek-V3        & \underline{81.3} \\

        \bottomrule
    \end{tabular*}
\end{table}

\paragraph{Zero-shot State of the Art on Spider.}
ZAS-SQL establishes a new zero-shot state of the art on Spider. With DeepSeek-V3, it reaches 87.2\% on Dev and 87.8\% on Test, surpassing C3-SQL by +5.4\%/+5.5\%. To isolate framework-level gains from backbone differences, we further evaluate ZAS-SQL with the same Qwen2.5-Coder-14B used by Alpha-SQL: ZAS-SQL obtains 87.2\% on Dev (+0.2\% over Alpha-SQL) and 88.6\% on Test, whereas Alpha-SQL reports no Test score. This controlled comparison confirms that the improvement stems from the ZAS-SQL framework itself rather than model choice, and the strong Dev/Test consistency demonstrates robust generalization.

\paragraph{Competitive with Few-shot and  Fine-tuning Methods.}
Without demonstrations, ZAS-SQL surpasses all few-shot GPT-4 baselines on Spider Dev except MCS-SQL, which relies on multi-path sampling. On Test, only MCS-SQL remains ahead. Against supervised methods, ZAS-SQL matches or exceeds the best fine-tuned result without task-specific training data.

\paragraph{Lightweight Models Rival Fine-tuned Larger Counterparts.}
ZAS-SQL with Qwen3-4B achieves 86.2\%/86.3\% on Spider Dev/Test, surpassing DTS-SQL and matching SENSE-SQL, despite using no training data. Models at or above the 4B level consistently reach 84\%-87\% regardless of model family, showing that the framework generalizes across diverse backbones. In contrast, Qwen2.5-3B-Instruct obtains only 69.7\%, revealing a capability floor below which rule-based constraints cannot fully compensate for insufficient language understanding.

\paragraph{Cross-Domain Transfer to UrbanPlan.}
As reported in Table~\ref{urbanplan_results}, ZAS-SQL with DeepSeek-V3 achieves 81.3\% on UrbanPlan. Inter-model variance is amplified: the gap between the strongest and weakest model is 36.0\% on UrbanPlan versus 17.5\% on Spider Dev, reflecting that domain-specific schema complexity magnifies differences in base-model reasoning capacity. Relative improvements over zero-shot baselines are consistently larger on UrbanPlan than on Spider across all models. This suggests that the marginal value of rule-driven constraints increases with query complexity: in the domain-specialized nested structures where vanilla prompting degrades most severely, explicit generation rules provide the greatest corrective effect.

\subsection{Ablation Study}\label{Ablation Study}
To verify the contribution of each module, we conduct a cumulative ablation on DeepSeek-V3 (Table~\ref{ablation}). Spider and UrbanPlan employ separate rule sets generated by the same distillation pipeline. The two sets are highly consistent along fundamental dimensions such as value matching, column selection, JOIN strategy, and sorting logic, yet each contains domain-specific rules: Spider emphasizes set operations and SQLite type handling, while UrbanPlan emphasizes window-function extrema, Common Table Expression multi-source merging, and temporal data processing. This divergence stems from structural differences in query patterns between the two datasets (e.g., UrbanPlan contains no set-operation queries but has a substantially higher proportion of nested subqueries than Spider).

\begin{table*}[t]
    \centering
    \caption{Cumulative ablation study on DeepSeek-V3. Modules are incrementally added to the zero-shot baseline; parenthetical values indicate per-step gains. Underlined entries mark the largest single-step improvement. Spider and UrbanPlan employ separate domain-adapted rule sets produced by the same distillation pipeline.}
    \label{ablation}
    \begin{tabular}{lccc}
        \toprule
        \textbf{Method} & \textbf{Dev} & \textbf{Test} & \textbf{UPlan} \\
        \midrule
        Zero-shot Baseline & 77.4 & 75.2 & 69.3 \\
        \quad w/ Schema Description & 79.5 (+2.1) & 77.3 (+2.1) & 72.9 (+3.6) \\
        \quad w/ Column Value Examples & 80.7 (+1.2) & 78.2 (+0.9) & 73.3 (+0.4) \\
        \quad w/ SQL Generation Rules & 85.2 (\underline{+4.5}) & 84.9 (\underline{+6.7}) & 78.7 (\underline{+5.4}) \\
        \quad w/ Step-by-Step Reasoning & 86.2 (+1.0) & 86.3 (+1.4) & 80.4 (+1.7) \\
        \quad w/ EGES & 87.2 (+1.0) & 87.8 (+1.5) & 81.3 (+0.9) \\
        \bottomrule
    \end{tabular}
\end{table*}

The contribution of each module is analyzed as follows.

\paragraph{Rules as the Dominant Factor.}
SQL generation rules yield the largest single-step gain across all three sets: +4.5\% on Dev, +6.7\% on Test, and +5.4\% on UrbanPlan, accounting for 45.0\%-53.2\% of the total cumulative improvement. The larger gain on Test correlates with lower baseline accuracy on Test, indicating that explicit constraints are most effective where initial error rates are highest.

\paragraph{Schema Enrichment with Domain-Dependent Asymmetry.}
Schema descriptions and column value examples jointly contribute +3.3\%/+3.0\%/+4.0\% on Dev/Test/UrbanPlan, but their relative balance differs by domain. Descriptions yield +3.6\% on UrbanPlan versus +2.1\% on Spider, reflecting greater semantic opacity of domain-specific column names. Conversely, value examples contribute only +0.4\% on UrbanPlan versus +1.2\% on Dev, consistent with the predominance of numerical predicates in UrbanPlan, where sample values offer limited disambiguation.

\paragraph{Reasoning and EGES Target Residual Errors.}
The five-step reasoning chain adds +1.0\%/+1.4\%/+1.7\% on Dev/Test/UrbanPlan. The larger gain on UrbanPlan aligns with its higher proportion of nested subqueries, where explicit intermediate decomposition prevents the model from conflating logical steps. EGES, operating atop an already strong pipeline, still delivers +1.0\%/+1.5\%/+0.9\%, confirming that execution-guided early stopping effectively reduces errors from sampling stochasticity, converting non-deterministic generation into reliable outputs.

\subsection{Multi-Model Zero-Shot Baseline Comparison}\label{Multi-Model Zero-Shot Baseline Comparison}
To assess the native capability boundaries of LLMs on Text-to-SQL, we benchmark fifteen models spanning the GPT-4o, DeepSeek-V3, GLM-5, Kimi-K2.5, Qwen2.5~\citep{46-hui2024qwen25codertechnicalreport,47-qwen2025qwen25technicalreport}, and Qwen3 families under a minimal zero-shot setting: DDL-only input, Chain-of-Thought (CoT) disabled, and $\tau = 0$. Three findings stand out: (1) open-source models now match or exceed GPT-4o, indicating that the proprietary advantage in structured code generation has largely disappeared; (2) complex queries remain a systematic bottleneck: nearly all models exceed 85\% on Easy queries, yet Extra accuracy drops to 34\%-72\%, a degradation of 20\%-50\%  within the same model; and (3) parameter scale yields diminishing returns: within the Qwen2.5 family, scaling from 3B to 14B yields +34.6\% on Dev, while the subsequent five-fold increase from 14B to 72B adds only +7.8\%. This sublinear scaling suggests that mid-sized models (14B-32B) augmented with ZAS-SQL offer a practical cost-performance sweet spot.

\section{Conclusion}\label{Conclusion}
This paper presents ZAS-SQL, a fully zero-shot framework addressing core failure modes of zero-shot Text-to-SQL through three complementary modules: knowledge-augmented schema representation, which supplements missing business semantics and value distributions in DDL via offline preprocessing; a structured reasoning framework, which embeds data-driven rules and a five-step reasoning process into the prompt to suppress over-complication; and EGES, which dynamically terminates sampling based on execution equivalence classes to achieve self-correction without iterative refinement.

Experiments demonstrate state-of-the-art zero-shot performance. On Spider, ZAS-SQL achieves 87.2\% and 88.6\% execution accuracy on Dev and Test, respectively, surpassing multiple few-shot and fine-tuning methods built upon GPT-4/4o. On UrbanPlan, it reaches 81.3\%, confirming cross-domain generalizability. Moreover, equipped with a 4B-parameter model, the framework surpasses zero-shot baselines of leading closed-source models, demonstrating strong model generality. Future work will extend the framework to benchmarks involving complex external knowledge such as BIRD, and explore automated iterative rule optimization to further reduce manual intervention.

\section*{Limitations}

This work has the following limitations.

\paragraph{Benchmark Coverage.} Experiments are primarily conducted on Spider, which uses SQLite as its execution engine. The robustness of the framework on enterprise-grade databases (e.g., Oracle, PostgreSQL) remains to be validated. Benchmarks such as BIRD introduce external domain knowledge and real-world noise, whereas the rule system of ZAS-SQL primarily targets SQL structural deviations and does not yet address knowledge-deficiency errors. Future work will explore joint distillation of structural and knowledge rules.

\paragraph{EGES Overhead and Parameter Sensitivity.} For complex queries with long execution times, the verification overhead of EGES remains non-negligible. The early stopping threshold $\theta$ is empirically determined and may require adaptive tuning to accommodate the difficulty distribution of a target domain.

\paragraph{Semantic Description Quality.} Automated semantic descriptions are inferred by the LLM without systematic human verification. For general-purpose databases, descriptions are typically reasonable; however, for highly specialized domains (e.g., medical or legal), accuracy may be insufficient and could introduce schema linking noise. Future work may incorporate lightweight manual spot-checking or automatic quality filtering based on downstream performance.

\paragraph{Rule Domain Transfer Cost.} Although the distillation pipeline generalizes across domains with minimal manual effort (Section~\ref{Core Rule Derivation}), the HITL abstraction step has not yet been fully automated. Fully automated iterative rule optimization remains a direction for future work.

\paragraph{Data Availability.} The UrbanPlan dataset is constructed from proprietary databases provided by our collaborating institution and cannot be publicly released due to data licensing restrictions. All experiments on Spider are fully reproducible.

\section*{Ethics Considerations}
The datasets and models used in this paper, and the implementation of the code and the resulting models, are not associated with any ethical concerns.


\bibliographystyle{acl_natbib}
\bibliography{custom}

\end{document}